\documentclass[journal,twoside,web]{ieeecolor}
\usepackage{tmi}
\usepackage{cite}
\usepackage{amsmath,amssymb,amsfonts}
\usepackage{algorithmic}
\usepackage{ulem}
\usepackage{graphicx}
\usepackage{textcomp}
\usepackage{multirow}
\def\BibTeX{{\rm B\kern-.05em{\sc i\kern-.025em b}\kern-.08em
    T\kern-.1667em\lower.7ex\hbox{E}\kern-.125emX}}
\begin{document}
\title{Topology-aware Pathological Consistency Matching for Weakly-Paired IHC Virtual Staining}
\author{Mingzhou Jiang,
        Jiaying Zhou, Nan Zeng, Mickaël Li,
        Qijie Tang, Chao He, Huazhu Fu,~\IEEEmembership{Senior Member, IEEE},
        and~Honghui He
\thanks{Manuscript received 6 January 2026. This work was supported by National Natural Science Foundation of China (NSFC) Grant No. 62335007. \textit{(Mingzhou Jiang and Jiaying Zhou  contributed equally to this work.) (Corresponding authors: Huazhu Fu and Honghui He)}}
\thanks{Mingzhou Jiang, Jiaying Zhou, Nan Zeng, Mickaël Li, Qijie Tang, and Honghui He are with Guangdong Research Center of Polarization Imaging and Measurement Engineering Technology, Shenzhen Key Laboratory for Minimal Invasive Medical Technologies, Institute of Biopharmaceutical and Health Engineering, Tsinghua Shenzhen International Graduate School,Tsinghua University, Shenzhen
518055, China (e-mail: 011122zmj@gmail.com; jiayingzhou2127@outlook.com; 
zengnan@sz.tsinghua.edu.cn; mickael.li@outlook.com; tqj24@mails.tsinghua.edu.cn;         he.honghui@sz.tsinghua.edu.cn).} 
\thanks{Chao He is with Department of Engineering Science, University of Oxford, Parks Road, Oxford OX1 3PJ, UK (e-mail: chao.he@eng.ox.ac.uk).}
\thanks{Huazhu Fu with the Institute of High Performance Computing (IHPC),
Agency for Science, Technology and Research (A*STAR), 138632, Singapore
(e-mail: hzfu@ieee.org).}
}
\maketitle

\begin{abstract}
Immunohistochemical (IHC) staining provides crucial molecular characterization of tissue samples and plays an indispensable role in the clinical examination and diagnosis of cancers. However, compared with the commonly used Hematoxylin and Eosin (H\&E) staining, IHC staining involves complex procedures and is both time-consuming and expensive, which limits its widespread clinical use. Virtual staining converts H\&E images to IHC images, offering a cost-effective alternative to clinical IHC staining. Nevertheless, using adjacent slides as ground truth often results in weakly-paired data with spatial misalignment and local deformations, hindering effective supervised learning. To address these challenges, we propose a novel topology-aware framework for H\&E-to-IHC virtual staining. Specifically, we introduce a Topology-aware Consistency Matching (TACM) mechanism that employs graph contrastive learning and topological perturbations to learn robust matching patterns despite spatial misalignments, ensuring structural consistency. 
Furthermore, we propose a Topology-constrained Pathological Matching (TCPM) mechanism that aligns pathological positive regions based on node importance to enhance pathological consistency. Extensive experiments on two benchmarks across four staining tasks demonstrate that our method outperforms state-of-the-art approaches, achieving superior generation quality with higher clinical relevance.

\end{abstract}


\section{Introduction}
\label{sec:introduction}
\IEEEPARstart{H}{istological} staining is an essential step in clinical tissue analysis and cancer diagnosis. Hematoxylin and Eosin (H\&E) staining is the most common and cost-effective staining technique used in pathology, where the hematoxylin primarily stains cell nuclei bluish-purple, and eosin stains cytoplasm pink \cite{fischer2008hematoxylin}. H\&E staining can provide a contrast between the nuclei and the extracellular matrix, highlighting morphological features of the tissue specimen \cite{bai2023deep}.

Despite the widespread use of H\&E staining, it sometimes fails to provide sufficient diagnostic information, such as specific protein expressions \cite{li2024virtual,fischer2008hematoxylin}. This limitation often renders H\&E staining unreliable for distinguishing among various cancer subtypes. Consequently, advanced techniques like Immunohistochemistry (IHC) are employed to target and visualize specific biomarkers that are not discernible with H\&E staining alone. IHC staining utilizes the principle of antigen-antibody binding to specifically label target cells \cite{yemelyanova2011immunohistochemical}. In a typical application, positive cells are stained brown by a chromogen, while a hematoxylin counterstain colors the nuclei of negative cells blue \cite{iqbal2014human}. IHC staining with different antibodies can reveal the organization and status of the tumor cells, which is crucial for distinguishing cancer subtypes and selecting personalized treatments. For instance, the expression of human epidermal growth factor receptor 2 (HER2) in breast tissue is scored from 0 to 3+, a result that directly guides the decision to use targeted therapy \cite{iqbal2014human, nitta2016assessment}.

Despite its advantages, IHC staining is more laborious, time-consuming, and expensive than routine H\&E staining. The long turnaround times (e.g., days to weeks) and the need for manual supervision by trained histotechnologists limit the widespread applicability of IHC staining, particularly in low-resource settings \cite{bai2023deep}. Therefore, in routine clinical practice, pathologists typically rely on the simpler H\&E staining for initial diagnosis. IHC staining is performed on adjacent tissue sections to obtain tumor-specific information only when further characterization is required. With the advent of digital pathology, virtual staining using deep learning has emerged as a promising and cost-effective alternative to conventional biochemical staining \cite{xu2025digital, jiang2025polarization}. This technique computationally transforms the source staining images into target staining images without laborious biochemical staining procedures. Developing deep learning models for H\&E-to-IHC virtual staining has been widely explored and validated, providing an effective solution for acquiring IHC staining images \cite{bai2023deep, li2024virtual, yang2025cross}. 
Previous methods can generally be categorized into fully supervised and unsupervised approaches. Pix2Pix \cite{isola2017image} and its variants are representative of the supervised methods, utilizing pixel-level supervision signals between cross-domain images to enhance image translation quality. However, in the task of virtual staining for pathological tissues, these methods often struggle to learn the mapping relationships of complex tissue structures. Unsupervised methods primarily utilize cycle-consistency constraints or contrastive learning to extract information from unpaired images, effectively preserving the details and structure of the input images. Adversarial loss is then employed to ensure style consistency between the stained image and the target domain image. A large body of work based on CycleGAN \cite{zhu2017unpaired} and CUT \cite{park2020contrastive} has progressively revolutionized the IHC staining task \cite{bai2023deep, pati2024accelerating}.

\begin{figure}[!t]
\centerline{\includegraphics[width=\columnwidth]{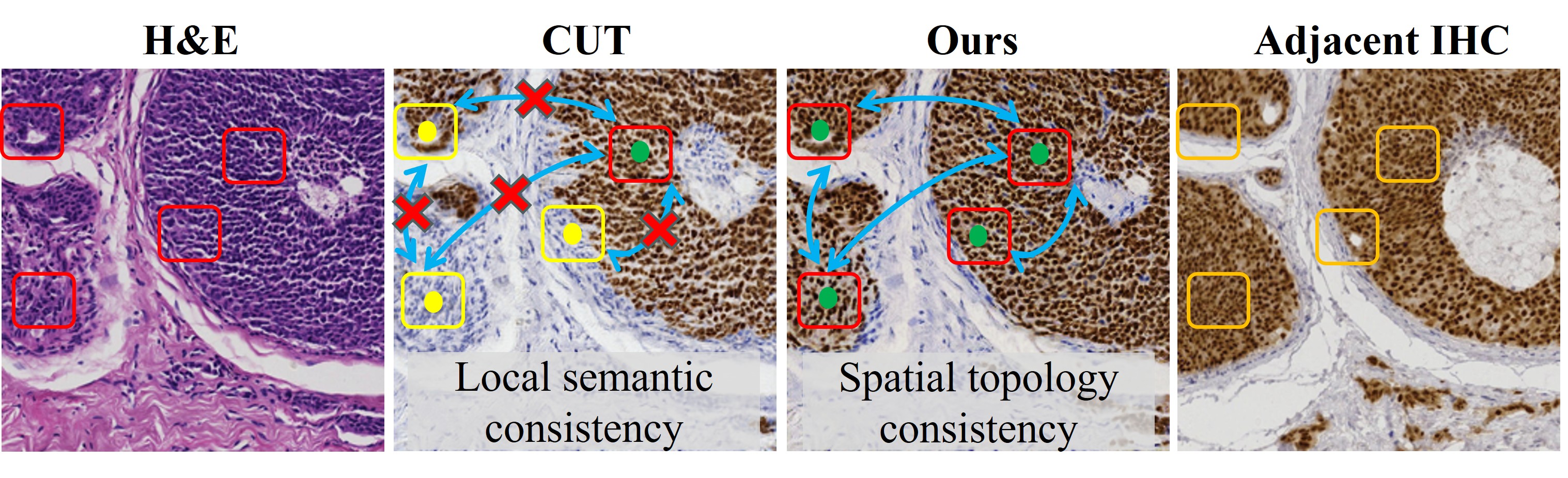}}
\caption{Spatial misalignment or local deformations between adjacent tissue sections can lead to misleading supervision signals. Previous methods, such as CUT\cite{park2020contrastive}, rely only on local semantic correspondence, resulting in inconsistent staining of regions with identical structures. In contrast, our method incorporates spatial topological matching to ensure consistent and coherent staining across corresponding regions.}
\label{fig1}
\end{figure}

The fidelity of H\&E-to-IHC virtual staining hinges on achieving both structural and pathological consistency. The former demands structural similarity between the generated IHC image and the source H\&E image, especially in regions sharing the same structures. The latter requires that the virtual stain accurately represents biomarker expression, thereby providing diagnostic information consistent with physical IHC slides for applications such as cancer subtyping. However, spatial misalignment or local deformations introduced during the staining process \cite{pati2024accelerating} often result in weakly-paired H\&E and IHC images.  When structural discrepancies exist between adjacent slides, the resulting supervision can lead to missing or inaccurate diagnostic information. Previous methods based on cycle-consistency constraints \cite{zhu2017unpaired, li2024virtual, lin2025virtual} or contrastive learning \cite{park2020contrastive, guan2025ot, wang2025oda} only focus on local semantic consistency between corresponding patches. Consequently, mismatched supervision information can lead to inconsistent staining in regions with identical structures, such as glands, as illustrated in Fig.\ref{fig1}. Therefore, explicitly modeling the spatial topological relationships among patch regions can further exploit valuable information and achieve structurally consistent staining.

In this paper, we propose the Topology-aware GAN (TA-GAN) for weakly-paired H\&E-IHC virtual staining. We use Graph Neural Networks (GNN) to learn the topological relationships between patches and introduce a Topology-aware Consistency Matching (TACM) mechanism to ensure staining structural consistency. Furthermore, a novel Topology-constrained Pathological Matching (TCPM) mechanism enhances positive region features using node importance scores, and accurately maps virtual IHC to real IHC based on correlation matching. Our main contributions are as follows:
\begin{itemize}
    \item We introduce a topology-aware method for IHC virtual staining. It consists of two information matching mechanisms: Topology-aware Consistency Matching (TACM) and Topology-constrained Pathological Matching (TCPM).
    \item TACM ensures staining structural consistency by leveraging topology-aware matching and topology-perturbation matching. It establishes topological relationships among pathological regions with the same structures.
    \item TCPM enhances positive topological nodes using node importance score and employs correlation matching to ensure staining pathological consistency.\footnote{\textcolor{blue}{Our code will be publicly available if the paper is accepted.}}
    \item Comprehensive evaluation on four H\&E-to-IHC virtual staining tasks across two datasets demonstrates the effectiveness of our method, surpassing the state-of-the-art (SOTA) IHC virtual staining methods. 
\end{itemize}

\section{Related Works}
\subsection{Image-to-Image translation}

The objective of image-to-image translation is to learn a mapping from a source domain to a target domain, transforming the style of an input image to that of the target domain while preserving its content \cite{richardson2021encoding, pang2021image}. Over the years, GAN-based approaches have proliferated for this task, with widespread applications in image translation across diverse domains, including natural and medical domains \cite{kaji2019overview}. Pix2Pix \cite{isola2017image}  established a conditional GAN paradigm using paired data, which effectively preserves information from the conditional input. This has made it a cornerstone of supervised image-to-image translation with paired data. To eliminate the need for paired data, CycleGAN \cite{zhu2017unpaired} revolutionized unsupervised image-to-image translation by introducing a cycle-consistency loss, which ensures that the generated image preserves the essential structure of the source image. Subsequent works, such as UNIT \cite{liu2017unsupervised}, MUNIT \cite{huang2018multimodal}, and DRIT \cite{lee2018diverse}, further advanced the field of unsupervised image translation. They explored the use of latent space encoding to handle unpaired data, which led to significant performance improvements. However, most of these methods assume cycle-consistency, which implies a strict bijection between the source and target domains. This assumption often fails in weakly-paired scenarios. To address this problem, CUT \cite{park2020contrastive} applies contrastive learning at the patch level to enforce correspondence between input and output images. It has since become a standard baseline for weakly-paired image translation. 



\subsection{Virtual IHC staining}
Virtual IHC staining has garnered increasing attention in recent years. Existing methods in this field rely primarily on the Generative Adversarial Network (GAN) paradigm. For example, Mercan et al. applied Conditional GAN and CycleGAN models to the task of translating H\&E images to the Phosphohistone-H3 (PHH3) IHC marker, demonstrating good consistency on breast tissue but noting the presence of sporadic artifacts in the brown staining \cite{mercan2020virtual}.Given the difficulty of obtaining precisely paired H\&E-IHC data, a significant body of work has utilized CycleGAN for unsupervised training \cite{liu2021unpaired, xie2022prostate, lahiani2020seamless}. To effectively leverage the supervision information in weakly-paired images, contrastive learning has emerged as a powerful technique. This approach works by mining information from positive and negative sample pairs at the patch level. The effectiveness of this paradigm has been validated on IHC virtual staining tasks for a variety of IHC biomarkers \cite{li2024virtual, zhang2024high, chen2024pathological, wang2025oda}. 
However, relying solely on pixel-wise or patch-level contrastive signals is often insufficient, as it may fail to accurately localize specific pathological positive regions in the virtually stained images.
Consequently, some recent works have begun to explore Optimal Transport (OT) theory to mine supervision information from weakly-paired data \cite{guan2025ot, guan2025supervised, peng2025usigan}. Although recent virtual IHC staining approaches have achieved promising progress, most existing methods primarily focus on pixel- or patch-level semantic correspondence while neglecting the spatial topological relationships among tissue regions. The lack of topology-aware modeling limits their ability to ensure structural alignment and pathological consistency, particularly under spatial misalignment and deformation between weakly-paired H\&E and IHC sections.

\begin{figure*}[t]
\centerline{\includegraphics[width=1.0\textwidth]{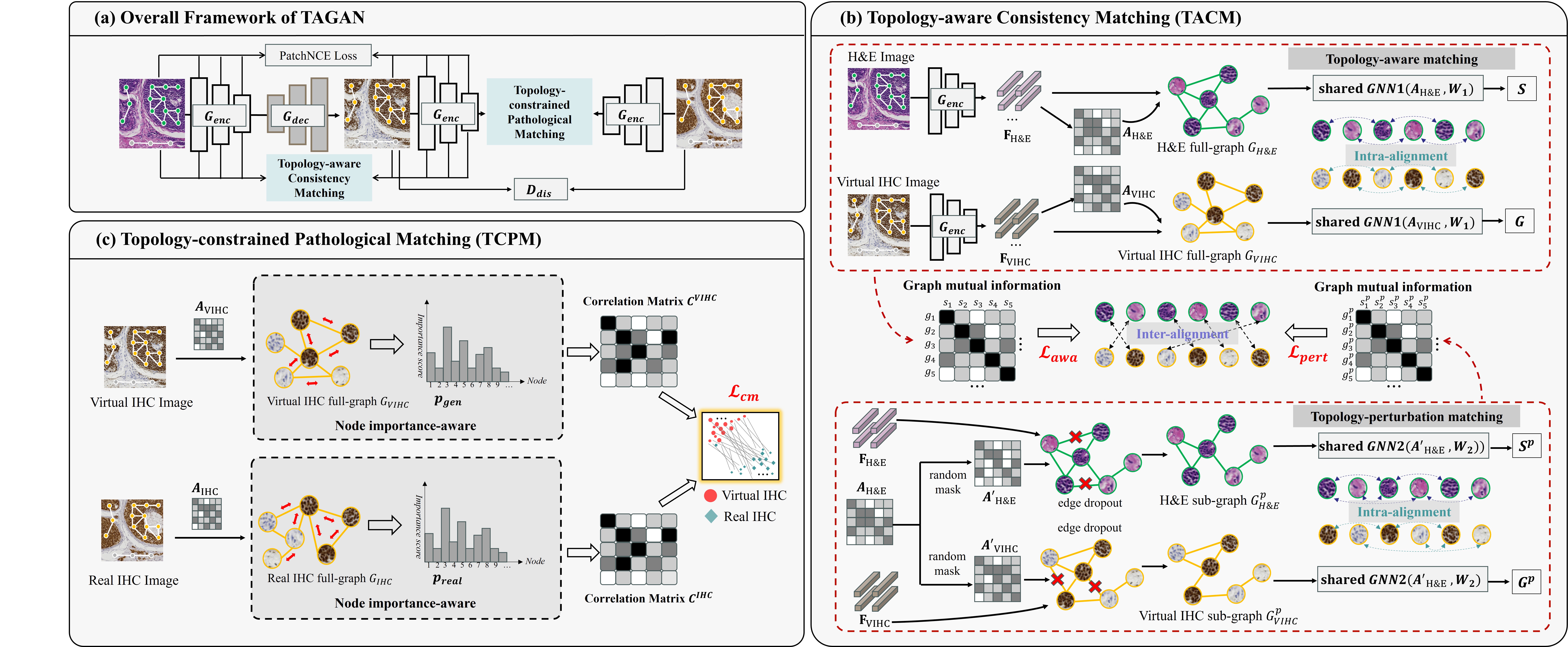}}
\caption{
(a) The overall architecture of TA-GAN. Following the general GAN paradigm, a generator takes an H\&E image as input and outputs an IHC stained image. The encoder of the generator is used to  extract features for the H\&E image, the generated virtual IHC image, and the real IHC image from adjacent slices. (b) The adjacency matrix is computed via feature cosine similarity. To ensure structural consistency, two GNN branches are employed: one aggregates information using their respective topologies, while the other applies H\&E topology with perturbations to the virtual IHC graph. Finally, a mutual information loss is calculated to align both branches. (c) Pathology Matching based on node importance. We compute node importance scores $p_{\text{gen}}$ and $p_{\text{real}}$
for the virtual and real IHC images, respectively, based on their connectivity relationships. These scores are then used to enhance the node features. Subsequently, we obtain correlation matrics $C^{\text{VIHC}}$ and $C^{\text{IHC}}$ via cosine dot product. Finally, $C^{\text{VIHC}}$ and $C^{\text{IHC}}$ are used to compute the correlation matching loss $\mathcal{L_{\text{cm}}}$.}
\label{fig2}
\end{figure*}

\section{Method}
The proposed TA-GAN paradigm employs a generator to map the input H\&E image to the IHC image domain. The generator’s encoder extracts features from the input H\&E stained image, the generated virtual IHC stained image, and the real IHC stained image. The topology-aware consistency matching (TACM) and the topology-constrained pathological matching (TCPM) mechanisms are used to ensure structural and pathological consistency, respectively. We introduce graph-based contrastive learning and correlation matching to guide the entire process, as shown in Fig.\ref{fig2}. 

\subsection{Topology-aware Consistency Matching (TACM)}

In clinical pathological practice, adjacent tissue sections stained with H\&E and IHC are used to represent distinct pathological features. However, due to the irreversibility of chemical staining, H\&E and IHC images often exhibit significant spatial misalignment or local deformations, leading to local feature mismatches. Previous contrastive learning-based methods rely on local semantic correspondences without considering the global topological relationships of pathological images, which results in inconsistent staining of identical pathological structures. To perform effective information matching from weakly-paired data, we propose the Topology-aware Consistency Matching (TACM) mechanism. As illustrated in Fig.\ref{fig2} (b), it consists of two branches: topology-aware matching and topology-perturbation matching. The former models the spatial dependencies between patches, while the latter robustly learns effective mappings from perturbed topological structures to ensure structural consistency during the staining process.

\subsubsection{Topology-aware matching} We treat the features of the source domain H\&E image, the generated virtual IHC image, and the real IHC image as nodes, and use a Graph Neural Network (GNN) to establish topological connections between patches. Specifically, given \( N \) features forming the feature set \( \mathbf{F} = [f_1, f_2, \dots, f_N]\in \mathbb{R}^{N \times D} \) representing the image, each feature is treated as a node in the graph. We construct the graph $G_{img} = \{ A, \mathbf{F} \}$, where \( A \) represents the adjacency matrix of the graph, which is computed from \( \mathbf{F} \). The adjacency matrix \( A \in \mathbb{R}^{N \times N} \) is computed using the dot product cosine similarity of node features. If the cosine similarity is greater than a preset threshold \( th \), it indicates that the two nodes are connected; otherwise, they are not. This ensures the correct modeling of the topological relationships between local patches. The calculation of \( A \) is as follows:

\begin{equation}
    A_{ij} := 
    \begin{cases} 
    1 & \text{if } \cos(f_i, f_j) \geq th \\
    0 & \text{if } \cos(f_i, f_j) < th
    \end{cases}
    \quad \text{for } i, j = 1, \dots, N
\end{equation}

Thus, by treating features as nodes and using the adjacency matrix to establish topological relationships, we construct the source domain H\&E image graph \( G_{\text{H\&E}} = \{ A_{\text{H\&E}}, \mathbf{F}_{\text{H\&E}} \} \) and the generated virtual IHC image graph \( G_{\text{VIHC}} = \{ A_{\text{VIHC}}, \mathbf{F}_{\text{VIHC}} \} \). 
Next, to leverage the inherent topological structure, we aim to propagate information across nodes with similar pathological morphologies as well as relevant background contexts.
To achieve this, a shared Topology Adaptive Graph Convolution Network \cite{du2017topology} is introduced to aggregate the graphs $G_{\text{H\&E}}$ and $G_{\text{VIHC}}$, which extracts graph representations to facilitate feature intra-alignment. Finally, we obtain the updated graph node features $S$ for the H\&E staining image and the graph node features 
$G$ for the generated IHC image. The calculation process is as follows:
\begin{equation}
\label{eq2}
    S = GNN1(A_{\text{H\&E}}, \mathbf{F}_{\text{H\&E}}) = \hat{A}_{\text{H\&E}}\mathbf{F}_{\text{H\&E}}W_1^{n}
\end{equation}
\begin{equation}
\label{eq3}
    \hat{A}_{\text{H\&E}} = D^{-1/2}_{\text{H\&E}} A_{\text{H\&E}} D^{-1/2}_{\text{H\&E}}
\end{equation}
\begin{equation}
\label{eq4}
    G = GNN1(A_{\text{VIHC}}, \mathbf{F}_{\text{VIHC}}) = \hat{A}_{\text{VIHC}}\mathbf{F}_{\text{VIHC}}W_1^{n}
\end{equation}
\begin{equation}
\label{eq5}
    \hat{A}_{\text{VIHC}} = D^{-1/2}_{\text{VIHC}} A_{\text{VIHC}} D^{-1/2}_{\text{VIHC}}
\end{equation}

\noindent
Where \( D_{\text{H\&E}} \) and \( D_{\text{VIHC}} \) are the degrees of the graphs \( G_{\text{H\&E}} \) and \( G_{\text{VIHC}} \), respectively. \( W_1^{n} \) represents the parameters of GNN1. \( n \) indicates the information from \( n \)-hops   neighboring nodes being aggregated to the current node. 

Through topology-aware matching, we constrain the graph node features of the H\&E image and the graph node features of the generated virtual IHC image to identify consistent pathological regions from the complex texture structures based on topological relationships. Finally, we align  the node mapping from H\&E to IHC by minimizing the mutual information loss, ensuring the consistency of stained structures in topologically related regions across the images. The mutual information loss is implemented using InfoNCE  \cite{oord2018representation} as follows:
\begin{equation}
    \mathcal{L_{\text{awa}}} = - \frac{1}{N} \sum_{i=1}^{N} \left[ \log \frac{\exp(s_i^\top g_i)}{\sum_{j=1}^{N} \exp(s_i^\top g_j)} \right]
\end{equation}
where $s_i$, $g_i$ are the i-th node features from $S$ and $G$.

\subsubsection{Topology-perturbation matching}
The generator mines information from weakly-paired H\&E and IHC images to learn the mapping relationship from H\&E to IHC. However, spatial misalignment between adjacent slices may perturb the topological structure. To address this, we propose topology-perturbation matching, which learns consistent representations under perturbed topological structures, resulting in more robust and stable mapping capabilities. 

Similarly, we construct the adjacency matrix $A_{\text{H\&E}}$ for the H\&E image using the method described earlier. Since our goal is to maintain strict structural consistency between the H\&E image and the generated virtual IHC image, the topological connectivity of patch features in the H\&E image should be similarly applicable to the generated IHC image. Therefore, we apply random masks to $A_{\text{H\&E}}$ with the same mask ratio  $m$ to remove edge nodes in the graph, resulting in two distinct adjacency matrices, $A_{\text{H\&E}}^{'}$ and $A_{\text{VIHC}}^{'}$, as shown in Fig.\ref{fig2} (b). Then, we construct  the perturbed graphs $G_{\text{H\&E}}^{p}$ and $G_{\text{VIHC}}^{p}$ using $A_{\text{H\&E}}^{'}$ and $A_{\text{VIHC}}^{'}$, respectively. Obviously, under lower mask ratios, $G_{\text{H\&E}}^{p}$ and $G_{\text{VIHC}}^{p}$ will maintain similar graph structures, but some regions' topological relationships will be removed. We aim to recover the original topological relationships from the perturbed graphs $G_{\text{H\&E}}^{p}$ and $G_{\text{VIHC}}^{p}$, which requires each node to fully leverage the information from all of its neighbors. Therefore, an $n$-hop GNN2 is used for topology-perturbation feature alignment, and its computation follows Eq.\ref{eq2}-\ref{eq5}. Finally, we obtain the graph node features $S^{p}$ and $G^{p}$ after restoring the topological relationships. Subsequently, we compute the mutual information loss to ensure the structural consistency of the mapping from the H\&E image to the virtual IHC image:

\begin{equation}
    \mathcal{L_{\text{pert}}} = - \frac{1}{N} \sum_{i=1}^{N} \left[ \log \frac{\exp((s_i^{p})^\top g_i^{p})}{\sum_{j=1}^{N} \exp((s_i^{p})^\top g_j^{p})} \right]
\end{equation}

where $s_i^{p}$, $g_i^{p}$ are the i-th node features from $S^{p}$ and $G^{p}$, respectively. Finally, the losses 
$\mathcal{L_{\text{awa}}}$ and 
$\mathcal{L_{\text{pert}}}$ constitute the structural consistency loss as follow:

\begin{equation}
    \mathcal{L_{\text{struc}}} = (\mathcal{L_{\text{awa}}} + \mathcal{L_{\text{pert}}})/2
\end{equation}

Topology-aware consistency matching leverages graph contrastive learning to learn structural consistency matching patterns from both the original graph connectivity and perturbed graph connectivity, thereby mining the information effectively.

\subsection{Topology-constrained Pathological Matching (TCPM)}
To further enhance the clinical validity of the generated IHC images, we employ topological constraints to mine critical pathological information from the weakly-paired real IHC images and generated virtual IHC images, as shown in Fig.\ref{fig2} (c). In clinical pathological diagnosis, different tissue regions contribute unequally to the overall diagnostic relevance. To capture this structural prior, we propose a node importance-aware feature enhancement mechanism to identify and emphasize diagnostically significant regions.

Given the feature set $\mathbf{F}\in \mathbb{R}^{N \times D}$, where $N$ denotes the number of patch nodes and $D$ represents the feature dimension, we first model the structural relationships among patches by constructing topological graph. The adjacency matrice $A$ is obtained following the procedure described above. In IHC images, pathological positive regions do not exist in isolation but exhibit intricate spatial relationships with their surrounding tissue contexts. This observation motivates a natural hypothesis: pathological positive nodes tend to exhibit higher connectivity, thereby acting as important nodes within the graph topology. To quantify the importance of each node within this graph, we employ the PageRank algorithm \cite{gleich2015pagerank}, which was originally designed to measure webpage importance based on link structure. Firstly, we compute the transition matrix $\mathbf{P}$ by $ \mathbf{P} = D^{-1}A$ ,
where $D$ is the degree of the graph. $\mathbf{P}$ represents the probability of transitioning from one node to its neighbors.  The node importance score vector $p\in \mathbb{R}^{N}$ is then computed via power iteration:

\begin{equation}
    p^{(t+1)} = \alpha \mathbf{P}^\top p^{(t)} + \frac{1-\alpha}{N}\mathbf{1}, \quad t = 0, 1, 2, \ldots
\end{equation}
where $p^{(0)} = \frac{1}{N}\mathbf{1}$ is initialized as a uniform distribution, $\alpha \in (0,1)$ is the damping factor controlling the probability of following graph edges versus random jumps, and $\mathbf{1}$ denotes an all-ones vector. The iteration proceeds until convergence $\|p^{(t+1)} - p^{(t)}\|_\infty < \epsilon$. The resulting scores $p$ characterizes the node-wise 
importance distribution, enabling effective discrimination between 
positive regions and surrounding normal tissue. 

We compute the importance scores $p_{\text{gen}}$ and $p_{\text{real}}$ 
for the generated virtual IHC graph \( G_{\text{VIHC}} = \{ A_{\text{VIHC}}, \mathbf{F}_{\text{VIHC}} \} \) and the real IHC 
graph \( G_{\text{IHC}} = \{ A_{\text{IHC}}, \mathbf{F}_{\text{IHC}} \} \), respectively. Subsequently, we utilize these 
importance scores to enhance the features of pathologically positive 
nodes. Specifically, we compute the element-wise product between the 
feature set and the importance scores, and add it to the original 
features as a residual connection. The calculation is as follows:
\begin{equation}
    \mathbf{F}_{\text{VIHC}} = \mathbf{F}_{\text{VIHC}} \odot p_{\text{gen}} + \mathbf{F}_{\text{VIHC}},   \quad \mathbf{F}_{\text{IHC}} = \mathbf{F}_{\text{IHC}} \odot p_{\text{real}} + \mathbf{F}_{\text{IHC}}
\end{equation}
where $\odot$ denotes element-wise multiplication with broadcasting. This design amplifies 
the representation of diagnostically important nodes while retaining 
the original feature information. Now we have two feature sets derived from the generated virtual IHC and real IHC respectively, in which the pathologically positive node features are emphasized. To learn the correct pathological distribution patterns from real IHC, we perform correlation matching. We compute the correlation matrix through self dot-product $C^{\text{VIHC}} = \mathbf{F}_{\text{VIHC}}\mathbf{F}_{\text{VIHC}}^\top$ and $C^{\text{IHC}} =\mathbf{F}_{\text{IHC}}\mathbf{F}_{\text{IHC}}^\top$, which reflects the connectivity patterns among node features. 

To ensure the pathological consistency between the generated and real IHC images, we propose a correlation matching loss, which guides that the real IHC and generated IHC share similar pathological region connectivity patterns. 
The formulation of the pathological correlation matching loss $\mathcal{L}_{cm}$ is defined as follows:

\begin{equation}
    \mathcal{L}_{cm} = \|C^{\text{VIHC}} - C^{\text{IHC}}\|
\end{equation}

Topology-constrained pathological matching utilizes the topological associations of patch nodes in real IHC images as supervision signals, which can effectively avoid signal misdirection caused by direct supervision from weakly-paired data. By combining TACM and TCPM,  TA-GAN matches the correct staining patterns from stained images of adjacent slices, which can help TA-GAN accurately generate IHC images.

\subsection{Training Details}
We use the CUT \cite{park2020contrastive} as the baseline for the network architecture, as shown in Fig.\ref{fig2} (a).
The loss function of TA-GAN includes adversarial loss 
$\mathcal{L_{\text{adv}}}$, PatchNCE loss $\mathcal{L_{\text{patchNCE}}}$, structural consistency loss $\mathcal{L_{\text{struc}}}$, and pathological correlation matching loss $\mathcal{L}_{cm}$. The  $\mathcal{L_{\text{adv}}}$ can be defined as:
\begin{equation}
    \mathcal{L_{\text{adv}}} = \mathbb{E}_{y \in Y} \log D_{dis}(y) + \mathbb{E}_{x \in X} \log(1 - D_{dis}(G(x)))
\end{equation}
where $\mathbb{E}$ denotes the expectation, $X$ and $Y$ denote the  source domain and target domain. $G(\cdot)$ represents Generator and $D_{dis}(\cdot)$ represents  Discriminator. The PatchNCE \cite{park2020contrastive} loss uses InfoNCE \cite{oord2018representation} to ensure content consistency during the image translation process. It is computed on the multi-scale features of the encoder and then averaged. The PatchNCE loss $\mathcal{L_{\text{patchNCE}}}$ can be formulated as:
\begin{equation}
   \mathcal{L_{\text{patchNCE}}}(X) = \mathbb{E}_{x \in X} \sum_{l=1}^{L} \sum_{s=1}^{S_l} \mathcal{L_{\text{infoNCE}}}(z, z^{+}, z^{-})
\end{equation}

where $L$ denotes the selected encoder layers, $S_l$
is the number of spatial locations in each
layer, $z$ represents the anchor embedding from output image.
$z^+$ is the positive embedding of the corresponding patch from the input image, while the $z^-$ are negative embeddings of the non-corresponding ones.

Finally, the total loss function is the weighted sum of the four individual losses, with all four losses jointly contributing to the update of the generator. The discriminator is updated using only the adversarial loss. Additionally, the structural consistency loss and pathological consistency loss are referenced from the PatchNCE loss and are similarly computed on multi-scale features from encoder, followed by averaging. The total loss is as follows:
\begin{equation}\label{eq16}
   \mathcal{L_{\text{total}}} = \mathcal{L_{\text{adv}}} + \mathcal{L_{\text{patchNCE}}} + \lambda_{1}\mathcal{L_{\text{struc}}} + \lambda_{2}\mathcal{L}_{cm}
\end{equation}

\section{Experiments and Results}
\subsection{Datasets}
We validated the effectiveness of our method on two externally sourced weakly-paired H\&E-to-IHC virtual staining datasets from different clinical settings. The first dataset is the MIST dataset \cite{li2023adaptive}, which contains weakly-paired H\&E and IHC stained patches from aligned breast tissue adjacent slices. There are four tasks, each with different IHC staining images (ER, PR, Ki67, HER2). Each task includes nearly 4,000 pairs of H\&E-IHC patches as the training set and 1,000 pairs of H\&E-IHC patches as the test set. All patches are of size  1,024 × 1,024 and non-overlapping.

The second dataset is the breast cancer immunohistochemical image generation challenge dataset BCI \cite{liu2022bci, zhu2023breast}. It consists of pairs of patches from adjacent slices, containing H\&E images and IHC staining targeting HER2. The dataset is also categorized into four classes (Class 0 to Class 3+) based on positive significance, from low to high. It was obtained by scanning pathological slices from over 600 breast cancer slices, and only well-aligned patches were retained after data filtering. The training set contains a total of 3,896 pairs of H\&E-IHC patches, and the test set contains 977 pairs of patches. All patches are of size 1,024 × 1,024 and non-overlapping.

\subsection{Implementation Details}
Due to memory limitations, all our experiments were conducted at a resolution of 256 × 256. We trained TA-GAN by resizing images to 256 × 256 resolution without any cropping. We used a 5-layer PatchGAN as the discriminator
and ResNet-6Blocks as the generator for TA-GAN. For the contrastive learning configurations, we followed the settings of CUT \cite{park2020contrastive},
e.g., 256 negative samples, temperature parameter $\tau$ = 0.07. Similarly, our loss is calculated at five scales on layers [0, 4, 8, 12, 16] of the generator. The corresponding thresholds $th$ for computing the adjacency matrix at these five scales are set to [0.5, 0.5, 0.1, 0.1, 0.1]. The adjacency matrix mask rate $m$ for calculating $\mathcal{L}_{\text{pert}}$ is set to 0.15. We use a 4-hops GNN for our model. The TA-GAN was trained for 100 epochs. During training, we used the Adam optimizer with a linear decay scheduler and an initial learning rate of 2 × $10^{-4}$. The hyperparameters
in Eq.\ref{eq16} were set as: $\lambda_{1}$ = 0.1 and $\lambda_{2}$ = 1.

Our method is implemented with PyTorch 2.1.2 and trained on a single NVIDIA RTX 3090 GPU (24 GB memory). All quantitative and qualitative experiment results reported in this study are based on the test set to evaluate the final performance of the proposed method. All reported results are averaged over multiple independent runs. 

\begin{table*}[htbp]
\centering
\caption{Quantitative comparison on the MIST and BCI dataset. Higher SSIM and PSNR, and lower FID and KID, indicate better generation quality. Since each subcategory (class0-class3+) in the BCI dataset has a limited number of samples, distribution-based metrics cannot be reliably computed.  Therefore, we report SSIM and PSNR for each subcategory, and evaluate FID and KID only on the full dataset. The best results are in bold and second in underline, respectively. }
\label{tab1}
\setlength{\tabcolsep}{3.5pt}
\renewcommand{\arraystretch}{1.1}
\begin{tabular}{c|cccc|cccc|cccc|cccc}
\hline
\multicolumn{17}{c}{\textbf{MIST dataset}} \\
\hline
\multirow{2}{*}{Method} & \multicolumn{4}{c|}{ER} & \multicolumn{4}{c|}{PR} & \multicolumn{4}{c|}{HER2} & \multicolumn{4}{c}{Ki67} \\
\cline{2-17}
 & SSIM & PSNR & FID & KID & SSIM & PSNR & FID & KID & SSIM & PSNR & FID & KID & SSIM & PSNR & FID & KID \\
\hline
Pix2Pix \cite{isola2017image} & 0.1151 & \underline{14.577} & 112.87 & 75.9 & 0.1189 & \underline{14.578} & 100.68 & 66.9 & 0.1047 & \textbf{14.751} & 127.40 & 67.5 & 0.1335 & 14.685 & 93.92 & 57.5 \\
CycleGAN \cite{zhu2017unpaired} & \textbf{0.1444} & 14.025 & 75.63 & 14.4 & 0.1244 & 14.230 & 62.23 & 10.7 & \textbf{0.1247} & 14.280 & 68.24 & 17.1 & \textbf{0.1544} & 14.771 & 64.24 & 20.5 \\
CUT \cite{park2020contrastive} & 0.1186 & 12.909 & 44.62 & 9.8 & 0.1262 & 13.659 & 53.72 & 24.5 & 0.0986 & 12.624 & 62.32 & 11.2 & 0.1343 & 13.797 & 43.06 & 11.7 \\
DDIB \cite{song2020denoising} & 0.0620 & 9.272 & 149.28 & 110.5 & 0.0575 & 9.209 & 179.58 & 140.1 & 0.0913 & 10.881 & 144.34 & 125.7 & 0.0584 & 9.450 & 185.26 & 159.3 \\
PyramidP2P \cite{liu2022bci} & 0.0987 & 14.111 & 107.19 & 62.2 & 0.1079 & 14.410 & 100.81 & 61.5 & 0.0947 & 14.529 & 126.15 & 96.0 & 0.1112 & 14.123 & 97.84 & 65.3 \\
ASP \cite{li2023adaptive} & 0.1269 & 14.334 & 41.13 & 5.9 & \textbf{0.1391} & \textbf{15.272} & 44.52 & 10.1 & 0.1096 & \underline{14.744} & 50.98 & 12.4 & 0.1306 & \textbf{15.543} & 50.79 & 19.6 \\
TDKStain \cite{peng2024advancing} & 0.1241 & 14.484 & 51.67 & 12.7 & 0.1216 & 14.560 & 51.05 & 13.5 & 0.1031 & 14.394 & 56.01 & 13.2 & 0.1305 & \underline{14.824} & 46.69 & 16.1 \\
PSPStain \cite{chen2024pathological} & 0.1268 & \textbf{14.644} & 91.81 & 37.3 & 0.1242 & 14.512 & 69.81 & 22.3 & 0.0947 & 13.829 & 85.54 & 26.1 & 0.1336 & 14.683 & 60.44 & 34.8 \\
USI-GAN \cite{peng2025usigan} & 0.1252 & 14.249 & \underline{37.68} & \underline{4.3} & 0.1223 & 14.356 & \underline{37.14} & \underline{4.7} & 0.1070 & 14.573 & \underline{46.18} & \underline{7.8} & 0.1312 & 14.475 & \underline{29.97} & \underline{3.9} \\
TA-GAN (Ours) & \underline{0.1314} & 13.877 & \textbf{31.48} & \textbf{1.6} & \underline{0.1322} & 14.144 & \textbf{31.29} & \textbf{1.6} & \underline{0.1105} & 13.832 & \textbf{38.47} & \textbf{3.2} & \underline{0.1356} & 14.441 & \textbf{25.77} & \textbf{1.5} \\
\hline
\multicolumn{17}{c}{\textbf{BCI dataset}} \\
\hline
\multirow{2}{*}{Method} & \multicolumn{3}{c|}{Class 0} & \multicolumn{3}{c|}{Class 1+} & \multicolumn{3}{c|}{Class 2+} & \multicolumn{3}{c|}{Class 3+} & \multicolumn{4}{c}{All} \\
\cline{2-17}
 & \multicolumn{3}{c|}{SSIM \hspace{2em} PSNR} & \multicolumn{3}{c|}{SSIM \hspace{2em} PSNR} & \multicolumn{3}{c|}{SSIM \hspace{2em} PSNR} & \multicolumn{3}{c|}{SSIM \hspace{2em} PSNR} & SSIM & PSNR & FID & KID \\
\hline
Pix2Pix \cite{isola2017image} & \multicolumn{3}{c|}{0.4032 \hspace{1.5em} 17.687} & \multicolumn{3}{c|}{0.3606 \hspace{1.5em} 19.015} & \multicolumn{3}{c|}{0.3474 \hspace{1.5em} \underline{19.589}} & \multicolumn{3}{c|}{0.3211 \hspace{1.5em} 17.472} & 0.3459 & \underline{18.825} & 113.36 & 32.3 \\
CycleGAN \cite{zhu2017unpaired} & \multicolumn{3}{c|}{\textbf{0.4507} \hspace{1.5em} 16.281} & \multicolumn{3}{c|}{0.3811 \hspace{1.5em} 17.220} & \multicolumn{3}{c|}{0.3529 \hspace{1.5em} 17.511} & \multicolumn{3}{c|}{\textbf{0.3687} \hspace{1.5em} 17.412} & 0.3676 & 17.367 & 57.05 & 19.5 \\
CUT \cite{park2020contrastive} & \multicolumn{3}{c|}{0.4165 \hspace{1.5em} \textbf{19.856}} & \multicolumn{3}{c|}{0.3406 \hspace{1.5em} 16.756} & \multicolumn{3}{c|}{0.3298 \hspace{1.5em} 15.154} & \multicolumn{3}{c|}{0.2980 \hspace{1.5em} 11.382} & 0.3355 & 14.778 & 67.86 & 19.4 \\
DDIB \cite{song2020denoising} & \multicolumn{3}{c|}{0.2353 \hspace{2em} 8.837} & \multicolumn{3}{c|}{0.2933 \hspace{1.5em} 11.309} & \multicolumn{3}{c|}{0.2747 \hspace{1.5em} 12.540} & \multicolumn{3}{c|}{0.2384 \hspace{1.5em} 13.144} & 0.2759 & 12.280 & 96.36 & 39.9 \\
PyramidP2P \cite{liu2022bci} & \multicolumn{3}{c|}{0.3580 \hspace{1.5em} 13.951} & \multicolumn{3}{c|}{0.3205 \hspace{1.5em} 18.004} & \multicolumn{3}{c|}{0.3002 \hspace{1.5em} 18.811} & \multicolumn{3}{c|}{0.2909 \hspace{1.5em} 17.620} & 0.3049 & 18.113 & 131.83 & 75.9 \\

ASP \cite{li2023adaptive} & \multicolumn{3}{c|}{0.3349 \hspace{1.5em} 15.960} & \multicolumn{3}{c|}{0.3066 \hspace{1.5em} 17.002} & \multicolumn{3}{c|}{0.2957 \hspace{1.5em} 17.355} & \multicolumn{3}{c|}{0.2627 \hspace{1.5em} 16.050} & 0.2911 & 16.859 & \underline{54.31} & \underline{13.1} \\
TDKStain \cite{peng2024advancing} & \multicolumn{3}{c|}{0.4116 \hspace{1.5em} \underline{18.451}} & \multicolumn{3}{c|}{0.3741 \hspace{1.5em} \textbf{21.299}} & \multicolumn{3}{c|}{0.3454 \hspace{1.5em} \textbf{20.439}} & \multicolumn{3}{c|}{0.3209 \hspace{1.5em} \underline{17.729}} & 0.3484 & \textbf{19.853} & 70.45 & 26.2 \\
PSPStain \cite{chen2024pathological} & \multicolumn{3}{c|}{0.4170 \hspace{1.5em} 17.948} & \multicolumn{3}{c|}{0.3861 \hspace{1.5em} \underline{20.839}} & \multicolumn{3}{c|}{0.3493 \hspace{1.5em} 18.766} & \multicolumn{3}{c|}{0.3242 \hspace{1.5em} 15.909} & 0.3542 & 18.478 & 90.58 & 31.9 \\
USI-GAN \cite{peng2025usigan} & \multicolumn{3}{c|}{0.4409 \hspace{1.5em} 14.049} & \multicolumn{3}{c|}{\textbf{0.4068} \hspace{1.5em} 17.525} & \multicolumn{3}{c|}{\textbf{0.3842} \hspace{1.5em} 18.077} & \multicolumn{3}{c|}{\underline{0.3610} \hspace{1.5em} \textbf{17.967}} & \textbf{0.3859} & 17.759 & 73.59 & 22.9 \\
TA-GAN (Ours)  & \multicolumn{3}{c|}{\underline{0.4411 }\hspace{1.5em} 16.520} & \multicolumn{3}{c|}{\underline{0.3947} \hspace{1.5em} 17.377} & \multicolumn{3}{c|}{\underline{0.3677 }\hspace{1.5em} 16.921} & \multicolumn{3}{c|}{0.3403 \hspace{1.5em} 14.990} & \underline{0.3807} & 16.567 & \textbf{40.55} & \textbf{5.6} \\
\hline
\end{tabular}
\end{table*}

\subsection{Evaluation Metrics}
In this work, we adopted four metrics to evaluate the performance of weakly-paired IHC virtual staining, including Structural Similarity Index Measure (SSIM) \cite{wang2004image}, Peak Signal-to-Noise Ratio (PSNR) \cite{hore2010image}, Fréchet Inception Distance (FID) \cite{heusel2017gans}, and Kernel Inception Distance (KID) \cite{binkowski2018demystifying}:
\subsubsection{SSIM} SSIM is used to measure the structural similarity between the generated image and the ground truth image. It evaluates image quality from three aspects of brightness, contrast, and structure. The SSIM is defined as:
\begin{equation}
    \text{SSIM} = \frac{(2\mu_x\mu_y + C_1)(2\sigma_{xy} + C_2)}{(\mu_x^2 + \mu_y^2 + C_1)(\sigma_x^2 + \sigma_y^2 + C_2)}
\end{equation}
where $\mu_x$ and $\mu_y$ are the means, $\sigma_x^2$ and $\sigma_y^2$ are the variances, $\sigma_{xy}$ is the covariance, and $C_1$, $C_2$ are constants to stabilize the division. Higher SSIM values indicate better structural similarity.

\subsubsection{PSNR} PSNR is used to measure the pixel-level similarity between the generated image and the ground truth image. It is defined based on the Mean Squared Error (MSE):
\begin{equation}
    \text{PSNR} = 10 \cdot \log_{10}\left(\frac{MAX^2}{\text{MSE}}\right)
\end{equation}
where $MAX$ is the maximum possible pixel value of the image, and and M SE is the mean square error between the images. Higher PSNR values indicate better image quality.

\subsubsection{FID} FID is used to measure the distributional distance between generated images and real images in the feature space. It computes the Fréchet distance between two multivariate Gaussians fitted to the features extracted from a pre-trained Inception network:
\begin{equation}
    \text{FID} = \|\mu_r - \mu_g\|^2 + \text{Tr}\left(\Sigma_r + \Sigma_g - 2(\Sigma_r\Sigma_g)^{1/2}\right)
\end{equation}
where $(\mu_r, \Sigma_r)$ and $(\mu_g, \Sigma_g)$ are the mean and covariance of the real and generated image features, respectively. Lower FID values indicate that the generated images are more similar to real images in terms of distribution.

\subsubsection{KID} 
KID is another metric for measuring the distributional similarity between generated and real images. Unlike FID, KID uses the squared Maximum Mean Discrepancy (MMD) with a polynomial kernel:
\begin{equation}
    \text{KID} = \mathbb{E}[k(x_r, x_r')] + \mathbb{E}[k(x_g, x_g')] - 2\mathbb{E}[k(x_r, x_g)]
\end{equation}
where $k(\cdot, \cdot)$ is a polynomial kernel function, and $x_r$, $x_g$ represent features of real and generated images. KID provides an unbiased estimator and is more reliable when the sample size is small. Following \cite{li2023adaptive}, we report KID values multiplied by $10^{3}$ for better readability.

It is worth noting that the H\&E images and IHC images from adjacent slices are weakly-paired, with potential spatial misalignment or deformation between them. SSIM and PSNR were originally designed for strictly pixel-aligned scenarios, and thus can only reflect approximate intensity and structural trends in our evaluation setting. Therefore, we prioritize FID and KID, which assess the distributional similarity between generated and real images and are more suitable for weakly-paired image translation tasks.

\subsection{Main Results}
We compared our TA-GAN model with state-of-the-art (SOTA) methods quantitatively and qualitatively based on the two datasets. The compared methods can be divided into the 2 types: 1) Image translation SOTA methods, including GAN-based Pix2Pix \cite{isola2017image}, CycleGAN \cite{zhu2017unpaired}, CUT \cite{park2020contrastive}, and diffusion-based DDIB \cite{song2020denoising}; 2) H\&E-to-IHC virtual staining SOTA methods, including PyramidP2P \cite{liu2022bci}, ASP \cite{li2023adaptive}, TDKStain \cite{peng2024advancing}, PSPStain \cite{chen2024pathological}, and USI-GAN \cite{peng2025usigan}. For fair comparison, all compared methods were either reproduced using official implementations or adopted from the original papers.

\subsubsection{Quantitative Comparison} 
\begin{figure*}[ht]
\centerline{\includegraphics[width=\textwidth]{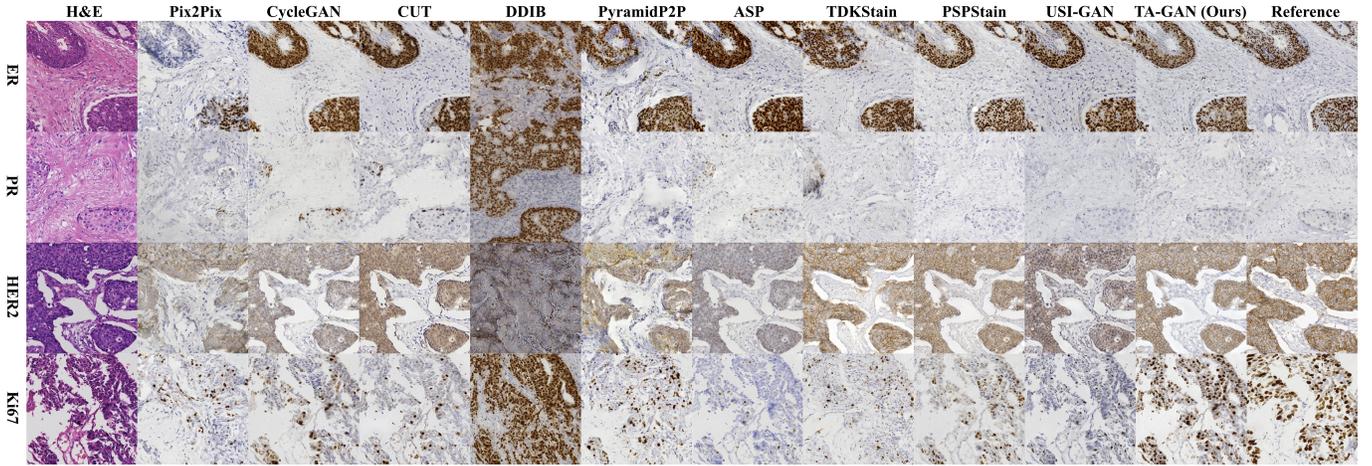}}
\caption{The quantitative
comparison of different state-of-the-art methods on the MIST dataset. The first column is the H\&E-stained images and the last columns is the adjacent slices real IHC images. Columns 2 to 11 are the images virtually stained by different methods.}
\label{fig3}
\end{figure*}

Table\ref{tab1} displays the quantitative experimental results on the MIST dataset and BCI dataset. Compared  with the SOTA methods, our approach TA-GAN achieves superior performance, setting a new state-of-the-art on these benchmarks. More specifically, our method significantly outperforms state-of-the-art image translation methods and virtual staining methods specifically designed for H\&E-to-IHC conversion on both FID and KID metrics. Since our method takes into account the structural consistency between input and output, it avoids structural drift caused by weakly-paired real IHC images, which leads to relatively modest performance on pixel-level metrics such as PSNR (calculated via MSE). However, as visualized in Fig.\ref{fig3}, our approach effectively preserves strong structural alignment compared with other methods. Fully supervised methods, such as Pix2Pix and Pyramid, directly compute the L1 loss between the virtually stained IHC images and the IHC images from adjacent tissue slices, which results in poor generation performance due to the misguidance from spatial misalignment or local deformations information.  Unsupervised methods such as CycleGAN  and CUT achieve better performance on FID and KID metrics. However, they fail to fully exploit the supervision signals from adjacent slices, yielding only suboptimal results. DDIB is a diffusion-based method that compresses the source domain into a latent space and then transforms it to the target domain. Due to the error deviation in the diffusion path, it produces poor generation quality. Specialized H\&E-to-IHC virtual staining methods, including ASP, TDKStain, PSPStain, and USI-GAN, employ contrastive learning based on inter-patch semantics and achieve relatively better generation quality. However, these methods neglect the spatial correlations among patches. In contrast, our proposed TA-GAN leverages TACM to establish matching patterns based on inter-patch spatial relationships, and utilizes TCPM to mine the spatial topological connections of pathological tissues from real IHC images, ultimately achieving optimal generation performance.

\subsubsection{Qualitative Comparison} We present the visual comparison results in Fig.\ref{fig3}. Compared with competing methods, the results generated by our method exhibit greater similarity to real IHC images in terms of color, structure, and texture details. Furthermore, our method addresses a previously mentioned issue by ensuring that the same tissue regions exhibit consistent staining patterns, thereby reducing the risk of erroneous staining or staining hallucinations. Qualitative results demonstrate the validity of our method.

\begin{figure}[t]
\centerline{\includegraphics[width=\columnwidth]{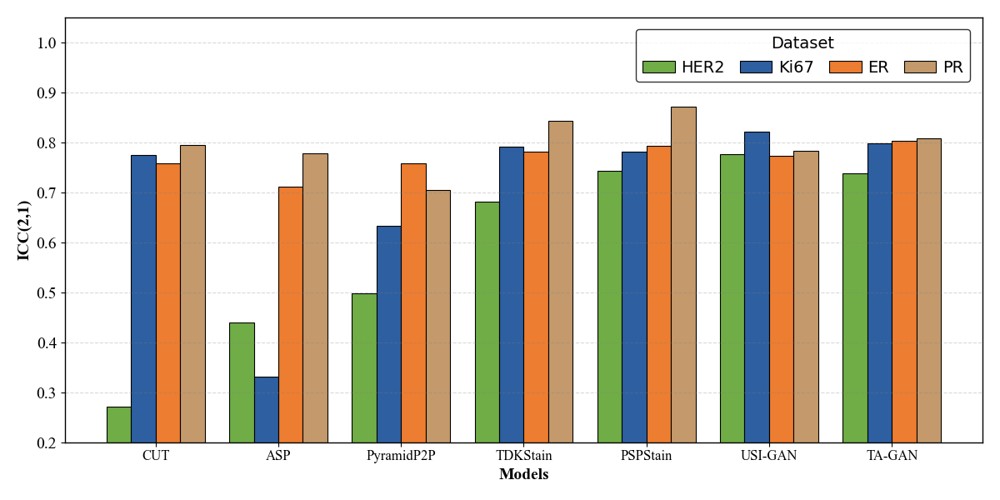}}
\caption{The Intraclass Correlation Coefficient (ICC) between the positive area ratios of the virtually stained IHC images and the real IHC images from adjacent tissue sections.}
\label{fig4}
\end{figure}

\begin{figure}[h]
\centerline{\includegraphics[width=\columnwidth]{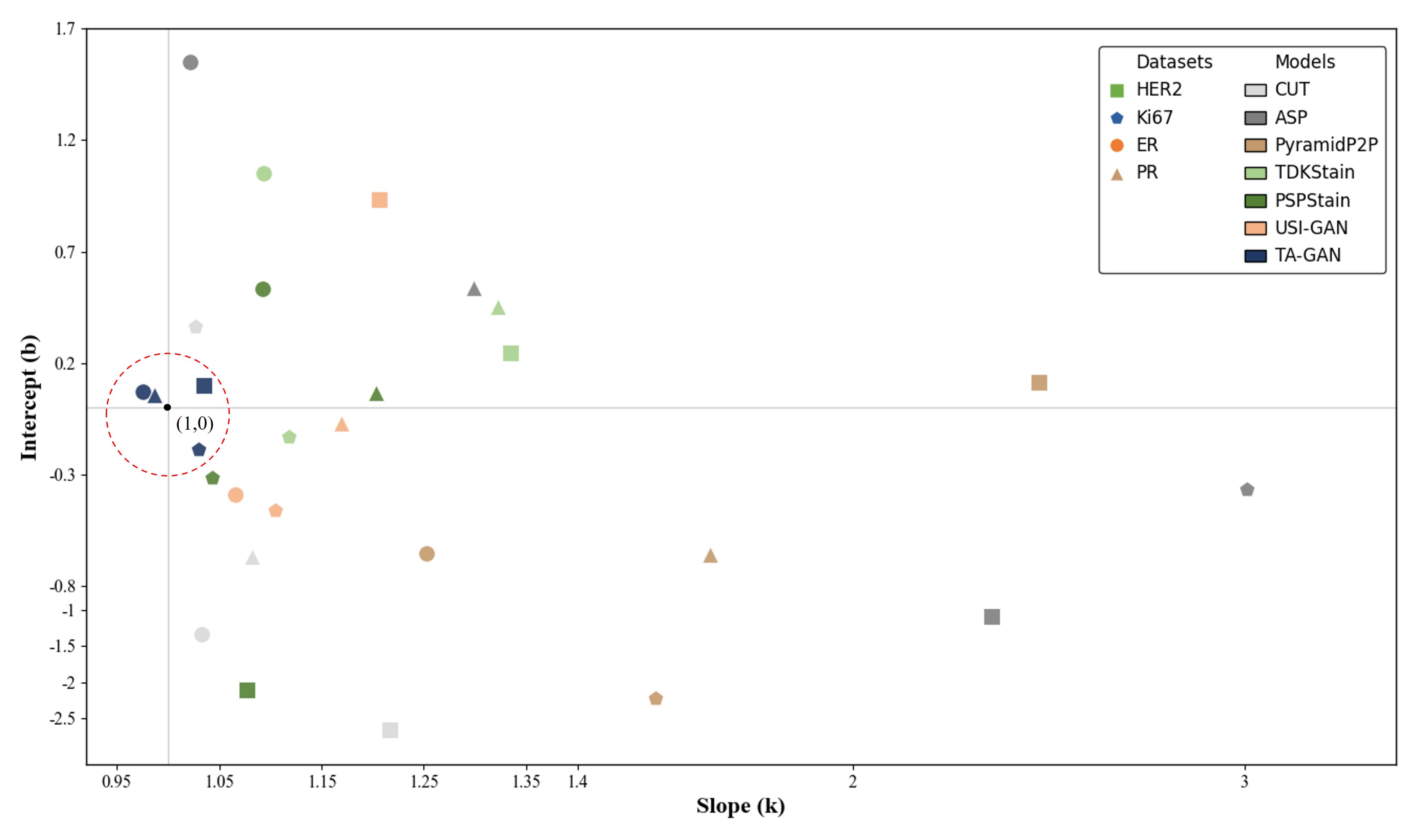}}
\caption{The trend equation derived from the scatter plot of positive area ratios uses the slope and intercept as its horizontal and vertical axes, respectively. The closer the position is to the ideal point (1, 0) (i.e., the line $y = x$), the stronger the agreement with the real IHC images.}
\label{fig5}
\end{figure}

\subsubsection{Pathological Consistency Assessment} 
Virtually stained IHC images should accurately reflect pathological grading, as this directly impacts the validity of clinical pathological diagnosis. Therefore, the accuracy of virtual staining can be demonstrated by evaluating the correlation of the positive area between the virtually stained IHC images and the real IHC images from adjacent tissue slices. We applied staining deconvolution to both the virtually stained IHC images and the paired IHC images from adjacent tissue slices, thereby separating each image into the hematoxylin (blue) and DAB (brown) channels. Subsequently, the percentage of positive area can be calculated from the DAB channel. Using the positive area ratios from both the virtually stained IHC images and the real IHC images, we computed the Interclass Correlation Coefficient (ICC) between them to evaluate the pathological consistency, as shown in Fig.\ref{fig4}. Figure\ref{fig4} presents the comparison results among our method TA-GAN, specialized IHC virtual staining methods, and the baseline CUT on MIST dataset. The results indicate that our method achieves high consistency scores. Although these scores are comparable to those of TDKStain, PSPStain, and USI-GAN, which employ explicit supervision based on the positive area of the DAB channel, the earlier results in Table\ref{tab1} and Fig.\ref{fig3} confirm that our method surpasses both in terms of generation quality. Furthermore, compared to the baseline CUT, we significantly improve pathological consistency, which verifies the effectiveness of our approach.
We further calculated regression equations reflecting the distribution trend based on scatter plots of the positive area, as shown in Fig.\ref{fig5}. Results closer to the line $y = x$ indicate a stronger correlation with the real IHC images. TA-GAN demonstrates higher correlation across all tasks, further validating the pathological consistency.

\begin{table}[h]
\centering
\setlength{\tabcolsep}{7pt}
\renewcommand{\arraystretch}{1.1}
\caption{Ablation study. Best results are in bold.}
\label{tab2}
\begin{tabular}{c|cccc}
\hline
Model & SSIM & PSNR & FID & KID \\
\hline
w/o all & 0.1186 & 12.909 & 44.62 & 9.8 \\
w TACM (w/o $\mathcal{L_{\text{pert}}}$) & 0.1110 & 12.534 & 40.58 & 9.7 \\
w TACM & 0.1301 & 13.764 & 39.08 & 5.9 \\
w TCPM & 0.1276 & 13.765 & 33.27 & 3.3 \\
TA-GAN (Ours) & \textbf{0.1314} & \textbf{13.877} & \textbf{31.48} & \textbf{1.6} \\
\hline
\end{tabular}
\end{table}
\subsubsection{Ablation Study} 
\begin{figure}[h]
\centerline{\includegraphics[width=\columnwidth]{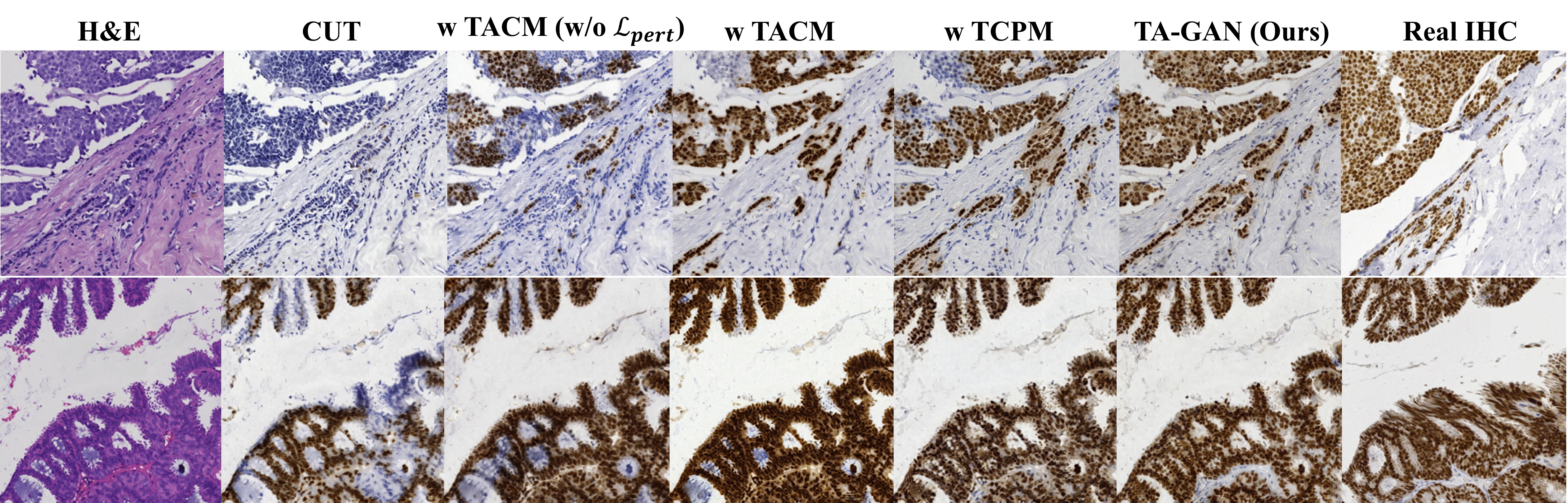}}
\caption{The visual comparison of the effects of TACM and TCPM.}
\label{fig6}
\end{figure}

We conduct comprehensive ablation studies on MIST dataset to demonstrate the effectiveness of our method. In Table\ref{tab2}, we demonstrate that the use of TACM and TCPM individually can both effectively improve the quality of generation, and the optimal results are achieved when the two are combined. This validates the effectiveness of the proposed TACM and TCPM. Furthermore, this is further verified by the qualitative visualization results in Fig.\ref{fig6}. TACM constrains the model to consider the spatial correlations of identical tissue regions, thereby improving the structural consistency of staining. TCPM helps the model mine effective pathological supervision information, improving the generation of pathological positive regions. When combining both modules, the images generated by the model show significant improvements in brightness, color, and structural texture, exhibiting greatest similarity to the real IHC images of adjacent slices. 

We use thresholds to determine the node connectivity relationships among patch features, and aggregate multi-hop neighbor information in the graph neural network to obtain node correlations. We conduct a comprehensive ablation analysis on the hyperparameters of the graph neural network in Table\ref{tab3}. As shown in Table\ref{tab3}, using larger thresholds for shallow-level features and smaller thresholds for high-level features can avoid the formation of fully-connected graphs and disconnected graphs, resulting in better generation performance. Additionally, increasing the number of hops for neighbor aggregation can gather more information, which also leads to improved generation results.
Based on this analysis, we choose 4-hops GNN and [0.5, 0.5, 0.1, 0.1, 0.1] as the value of threshold $th$.

\begin{table}[ht]
\centering
\setlength{\tabcolsep}{5pt}
\renewcommand{\arraystretch}{1.1}
\caption{Ablation study about the hyperparameters $n$-hops and threshold $th$. Threshold $th$ values correspond to feature layers [0,4,8,12,16]. Best results are in bold.}
\label{tab3}
\begin{tabular}{c|c|cccc}
\hline
$n$-hops & Threshold $th$ & SSIM & PSNR & FID & KID \\
\hline
2 & [0.1, 0.1, 0.1, 0.1, 0.1] & 0.1268 & 13.568 & 58.96 & 15.5 \\
3 & [0.1, 0.1, 0.1, 0.1, 0.1] & 0.1245 & 13.716 & 42.84 & 14.1 \\
4 & [0.1, 0.1, 0.1, 0.1, 0.1] & 0.1178 & 13.496 & 42.46 & 10.3 \\
2 & [0.5, 0.5, 0.1, 0.1, 0.1] & 0.1254 & 13.617 & 49.90 & 19.9 \\
3 & [0.5, 0.5, 0.1, 0.1, 0.1] & 0.1253 & \textbf{14.105} & 39.58 & 8.4 \\
4 & [0.5, 0.5, 0.1, 0.1, 0.1] & 0.1301 & 13.764 & \textbf{39.08} & \textbf{5.9} \\
4 & [0.5, 0.5, 0.5, 0.5, 0.5] & \textbf{0.1413} & 13.441 & 49.66 & 9.1 \\
\hline
\end{tabular}
\end{table}

To investigate the impact of different mask rates $m$ in Topology-perturbation matching, we conduct an ablation study with various values of mask rates $m$. As shown in Table\ref{tab4}, the optimal performance is achieved when the mask rate is set to 0.15. A lower mask rate results in the H\&E image features and generated IHC image features sharing the same adjacency matrix, whereas a higher mask rate destroys the original feature correlations. Both scenarios lead to suboptimal performance. 

In addition, to investigate the impact of different weight
combinations of $\lambda_{1}$ and $\lambda_{2}$, we conducted an ablation study with various
values of $\lambda_{1}$ and $\lambda_{2}$, as shown in Table\ref{tab5}. TA-GAN achieves the best performance when the loss weights are set to $\lambda_{1}=0.1$
$\lambda_{2}=1$, and also achieves high performance across various settings. This demonstrates that our method exhibits good training stability without requiring additional regularization constraints to aid convergence.

\begin{table}[htbp]
\centering
\setlength{\tabcolsep}{7pt}
\renewcommand{\arraystretch}{1.1}
\caption{Effect of mask ratio $m$. Best results are in bold.}
\label{tab4}
\begin{tabular}{c|cccc}
\hline
mask ratio $m$ & SSIM & PSNR & FID & KID \\
\hline
0.05 & 0.1144 & 13.159 & 51.77 & 17.6 \\
0.1 & 0.1291 & 13.715 & 40.81 & 9.3 \\
0.15 & 0.1301 & 13.764 & \textbf{39.08} & \textbf{5.9} \\
0.2 & \textbf{0.1342} & \textbf{14.320} & 44.91 & 12.1 \\
\hline
\end{tabular}
\end{table}

\begin{table}[h]
\centering
\setlength{\tabcolsep}{7pt}
\renewcommand{\arraystretch}{1.1}
\caption{Effect of different loss weight for $\lambda_{1}$ and $\lambda_{2}$. Best results are in bold.}
\label{tab5}
\begin{tabular}{c|cccc}
\hline
Loss weight & SSIM & PSNR & FID & KID \\
\hline
$\lambda_{1}=0.1$, $\lambda_{2}=0.01$ & 0.1246 & 13.522 & 32.88 & 2.7 \\
$\lambda_{1}=0.1$, $\lambda_{2}=0.1$ & 0.1251 & 13.581 & 32.91 & 2.2 \\
$\lambda_{1}=0.1$, $\lambda_{2}=1$ & \textbf{0.1314} & \textbf{13.877} & \textbf{31.48} & \textbf{1.6} \\
$\lambda_{1}=0.1$, $\lambda_{2}=10$ & 0.1304 & 13.829 & 31.91 & 2.0 \\
$\lambda_{1}=0.1$, $\lambda_{2}=100$ & 0.1254 & 13.854 & 34.48 & 2.4 \\
\hline
\end{tabular}
\end{table}

\begin{figure}[h]
\centerline{\includegraphics[width=0.8\columnwidth]{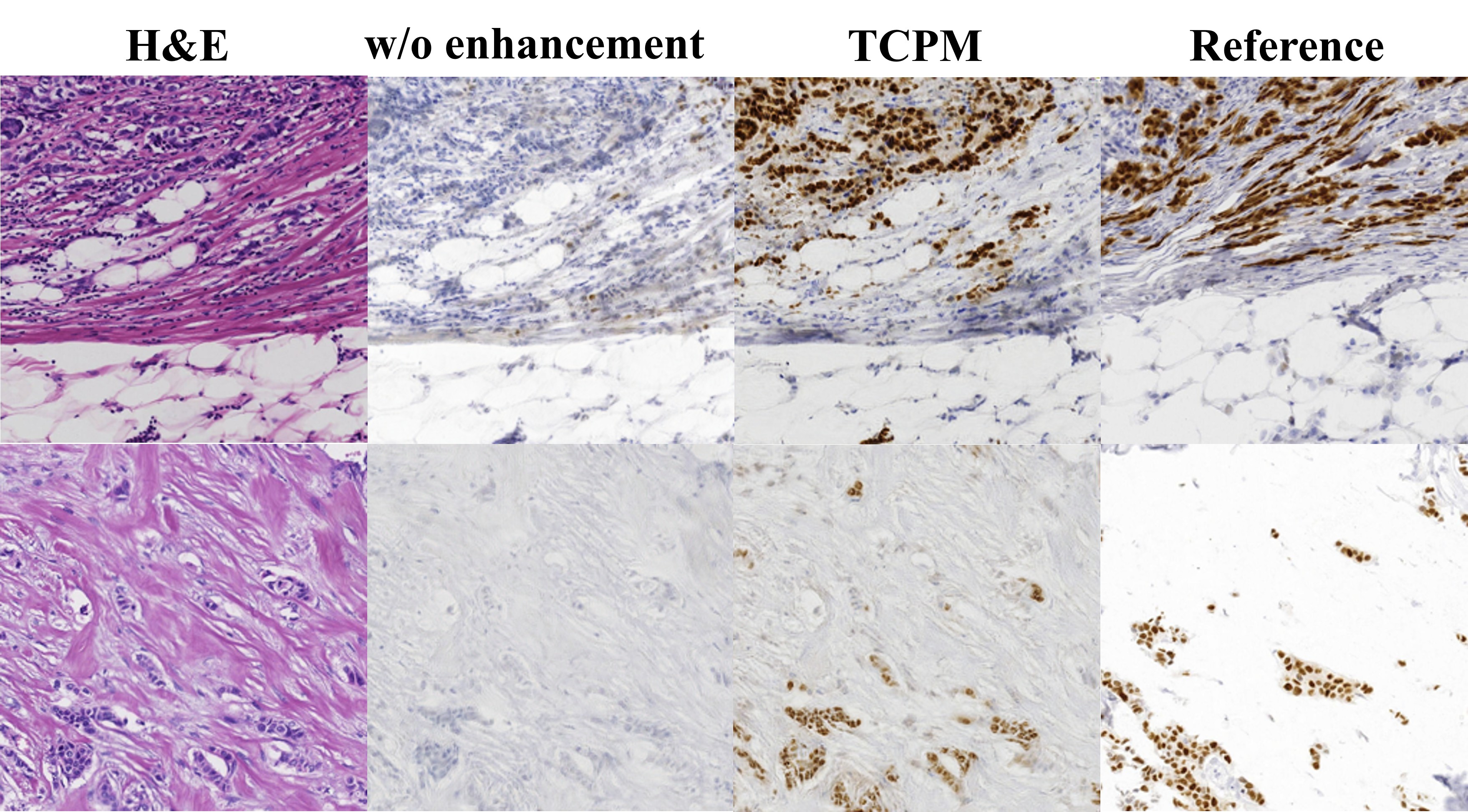}}
\caption{The visual comparison of the effects of node importance-
aware feature enhancement in TCPM.}
\label{fig7}
\end{figure}

Furthermore, we conduct an ablation study to investigate the effect of the node importance-
aware feature enhancement in TCPM, as shown in Table\ref{tab6}. Node enhancement in positive regions, guided by topological relationships among node features, yields significant gains in generation quality. This evidences the validity of the TCPM design. The visual comparison in Fig.\ref{fig7} further demonstrates that TCPM can more accurately identify pathological positive regions, thereby ensuring pathological consistency.

\begin{table}[htbp]
\centering
\setlength{\tabcolsep}{7pt}
\renewcommand{\arraystretch}{1.1}
\caption{effect of the node importance-
aware feature enhancement in TCPM.}
\label{tab6}
\begin{tabular}{c|cccc}
\hline
Method $m$ & SSIM & PSNR & FID & KID \\
\hline
w/o enhancement & 0.1310 & 13.611 & 35.21 & 3.6 \\
TCPM (Ours) & 0.1276 & 13.765 & 33.27 & 3.3 \\
\hline
\end{tabular}
\end{table}

\section{Conclusion}
In this study, we proposed a topology-aware weakly supervised framework, namely TA-GAN, for weakly-paired H\&E-to-IHC virtual staining. Previous contrastive learning-based IHC virtual staining methods only consider the semantic associations among patches while neglecting the spatial correlations of tissue regions. To address this problem, here we proposed a Topology-aware Consistency Matching (TACM) mechanism, which performs graph contrastive learning at both the normal feature topology graph and perturbed graph levels, ensuring the correct mining of supervision signals under the weakly-paired data scenario. We also developed a Topology-constrained Pathology Matching (TCPM) mechanism, which obtains importance scores of patch nodes based on the connectivity of topology graphs to enhance positive region features, and employs correlation matching to ensure that the generated IHC images and real IHC images exhibit similar topological connectivity patterns. Comprehensive evaluations on four H\&E-to-IHC staining tasks across two datasets validated the effectiveness of our proposed method. Both quantitative and qualitative results demonstrated that our approach outperforms previous state-of-the-art methods, achieving superior staining consistency and pathological consistency. 

Although TA-GAN achieved superior performance, there is still area for improvement. TA-GAN requires training a separate generator for each IHC staining task without considering the common features across different IHC virtual staining tasks. Multi-domain transferring strategies are worth exploring in future work. Additionally, graph neural networks are sensitive to hyperparameters. The current results of TA-GAN are obtained through manual hyperparameter tuning, and automatic hyperparameter search could be considered in the future to achieve better results.


\bibliographystyle{IEEEtran}
\bibliography{reference}
\end{document}